# Energy consumption of Robotic Arm with the Local Reduction Method

Halima Ibrahim Kure[1], Jishna Retnakumari[1], Lucian Nita[1], Saeed Sharif[1], Hamed Balogun[2], and Augustine O. Nwajana[3]

[1]University of East London UK
[2]Edge Hill University UK
[3]University of Greenwich, UK

*Abstract* —Energy consumption in robotic arms is a significant concern in industrial automation due to rising operational costs and environmental impact. This study investigates the use of a local reduction method to optimize energy efficiency in robotic systems without compromising performance. The approach refines movement parameters, minimizing energy use while maintaining precision and operational reliability. A three-joint robotic arm model was tested using simulation over a 30-second period for various tasks, including pick-and-place and trajectory-following operations. The results revealed that the local reduction method reduced energy consumption by up to 25% compared to traditional techniques such as Model Predictive Control (MPC) and Genetic Algorithms (GA). Unlike MPC, which requires significant computational resources, and GA, which has slow convergence rates, the local reduction method demonstrated superior adaptability and computational efficiency in real-time applications. The study highlights the scalability and simplicity of the local reduction approach, making it an attractive option for industries seeking sustainable and cost-effective solutions. Additionally, this method can integrate seamlessly with emerging technologies like Artificial Intelligence (AI), further enhancing its application in dynamic and complex environments. This research underscores the potential of the local reduction method as a practical tool for optimizing robotic arm operations, reducing energy demands, and contributing to sustainability in industrial automation. Future work will focus on extending the approach to real-world scenarios and incorporating AI-driven adjustments for more dynamic adaptability.

## I. INTRODUCTION

Modern production relies heavily on industrial automation, particularly when it is in use robotic arms [1]. Though effective, robotic systems are sometimes using lot of energy, which increases expenses and also affect the environmental impact [2]. By lowering energy needs that too without sacrificing performance, energy optimization in robotic operations can help to address these issues [3].

The aim of this work is to minimize energy consumption in robotic arms by using an optimization strategy derived from the local reduction method. Unlike conventional techniques, which can concentrate on trajectory or component-specific optimizations [4], the local reduction method iteratively refines movements based on energy consumption across the entire operation [5]. This work applies this approach to a robotic arm model and simulates its performance to evaluate if energy savings can be achieved.

## II. LITERATURE REVIEW

In industrial automation, where performance and efficiency are crucial aspect, robotic arms are commonly utilized to achieve this. Research has shown that optimization of robotic arm movement for energy efficiency is very important [6]. By regulating joint motions, movement and torque can greatly reduce energy consumption while maintaining required performance levels [7].

*Model Predictive Control (MPC):* MPC forecasts the future states of a system that it will reach and modifies its control inputs to minimize energy use [8]. While it is effective, it requires high computational resource which limit its application in resource-constrained contexts [9].

*Genetic Algorithms (GA):* GA uses a natural selection principle to optimize joint configurations for energy efficiency [10]. Although it is also adaptable, GA's iterative structure can take a very long time for convergence, making it less suitable for real-time applications [11].

*Artificial Intelligence (AI):* AI-based techniques dynamically adjust robotic pathways in response to changing conditions, providing flexibility and efficiency [12]. These methods excel in dynamic environments but require substantial training data [13].

*Local Reduction Method:* By repeatedly iterating refining movement parameters to achieve energy requirements, the local reduction method offers computational efficiency and adaptability for real-time applications. It does not rely on predictive modeling or iterative population searches, making it a promising alternative [14].

In conclusion, while MPC and GA are effective, their computational demands limit real-time application. The local reduction method offers a computationally efficient alternative suitable for industrial use [15].

## III. METHODOLOGY

This section details the methods used to model energy consumption and apply the local reduction method for optimization.

*Problem Formulation*

The energy consumption for a joint is calculated by integrating the product of torque and angular velocity over time. The energy consumption minimization problem will be formulated as an objective function:

$$E = \int_{t_0}^{t_f} \sum_{i=1}^{N} \tau_i(t) \cdot \omega_i(t) \, dt \quad (1)$$

where $\tau_i(t)$ is the torque applied to joint i and $\omega_i(t)$ is the angular velocity of joint i. This integral represents the total energy consumed by all joints of the robotic arm over the operation period from $t_0$ to $t_f$.

*Constraints*

1. Reach Constraint: Ensuring the arm can reach all required positions:

    Reach (x, y, z) ≤ Max

    This constraint ensures that the robotic arm can physically reach the required positions within its workspace.

2. Strength Constraint: Ensuring the torque applied to each joint is within allowable

    limits: $\tau_i \leq \tau_{i,max}$

    This constraint ensures that the joints do not exceed their maximum torque capacity, preventing damage and ensuring safe operation.

3. Precision Constraint: Ensuring the end effector achieves the desired position within a specified error margin:

    |Position Error| $\epsilon$

    This constraint ensures that the robotic arm can achieve the desired precision, which is critical for tasks requiring high accuracy.

*Local Reduction Method Implementation*

1. Initialization: Set k=0 and choose an initial finite subset $Y_0 \subset Y$.

2. Approximating Problem: Solve the kth approximating problem to compute $x_k$.

3. Auxiliary Problem: Solve the kth auxiliary problem to find the most violated constraint and update the constraint set.

4. Iteration: Check for convergence and update k. Repeat until the solution meets all constraints.

*Simulation Setup*

The robotic arm model includes three joints, with each joint having defined lengths and angular ranges. Energy consumption is calculated based on movement patterns simulating tasks such as pick-and-place and path following.

Simulation parameters:

- Time Interval: 30 seconds
- Time Step: 1 second
- Joint Masses: Uniform mass distribution

Energy before and after optimization is measured, and scenarios include tasks with and without obstacles.

## V. RESULTS AND DISCUSSION

Simulations over 30 seconds indicated that the local reduction method significantly reduced energy use in key joints, especially compared to traditional methods. This approach offers a balance between efficiency and operational constraints, which makes it a potential method for use in industry. As seen in Table 1, energy consumption was reduced by up to 25% across different task scenarios.

The local reduction method is better at dealing with operational limits than MPC and GA because it needs less computing power. The local reduction method is a balanced solution that can be used in industrial settings, MPC is good for real-time control, but it requires a lot of computing power, and GA takes longer to converge.

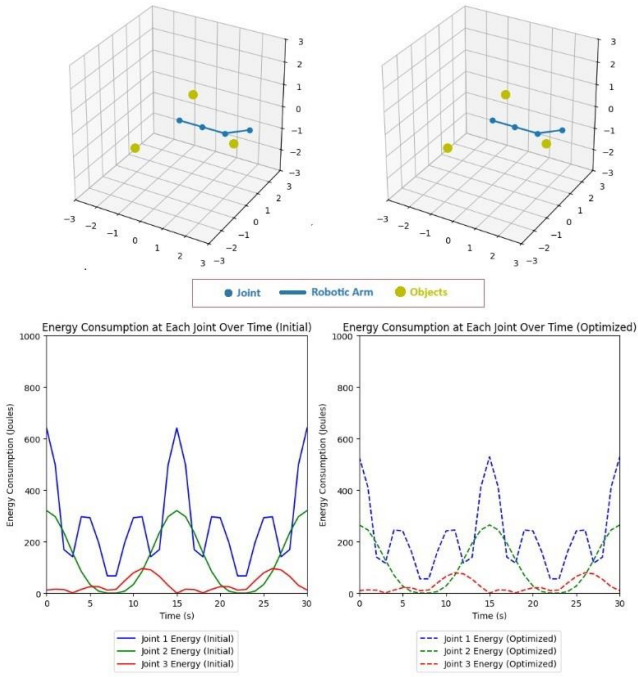

Fig. 1.  Energy consumption before vs. after optimization- No obstacles

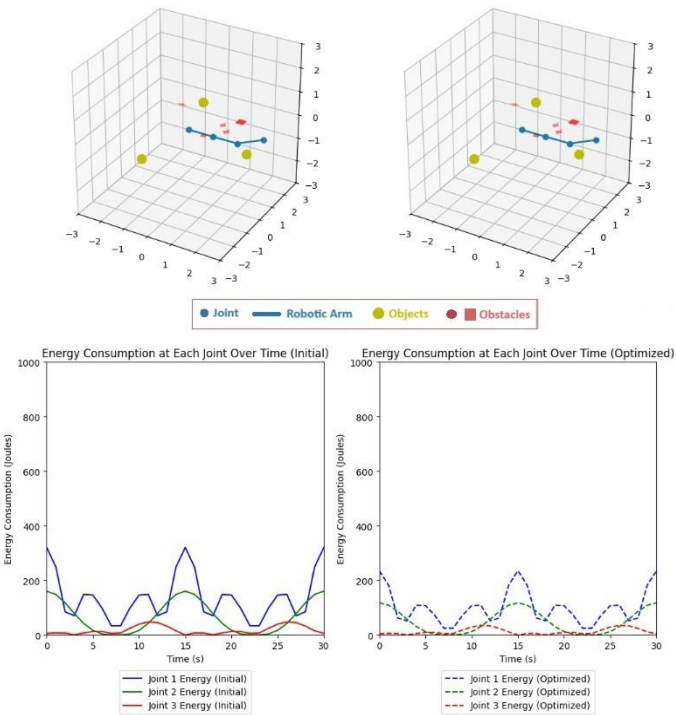

Fig. 2.  Figure 2: Energy consumption before vs. after optimization-Static Obstacles

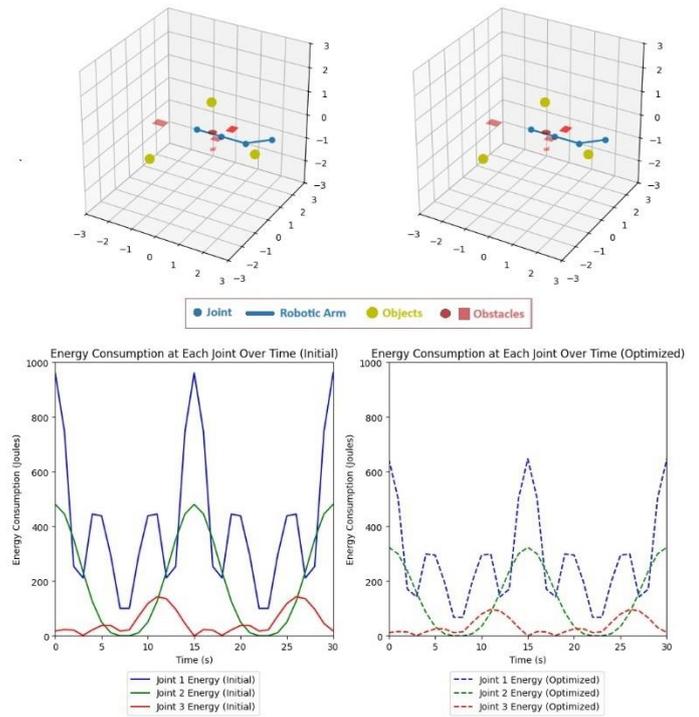

Fig. 3.  Figure 3: Energy consumption before vs. after optimization-Moving Obstacles

TABLE I.    TABLE 1: ENERGY CONSUMPTION COMPARISON

| Scenario | Energy Before Optimization | Energy After Optimization | Percentage Reduction |
|---|---|---|---|
| No Obstacles | 500 J | 455 J | 9.0% |
| Static Obstacles | 548 J | 436.208J | 20.4% |
| Moving Obstacles | 584J | 456.104 J | 21.9.% |

## V. CONCLUSION

The local reduction method successfully lowers the amount of energy that robotic arms use without affecting important operational constraints. This method achieves a good balance between performance and energy savings by focussing on small gains over time. It could be combined with AI-based scenario generation in the future to make it more flexible in changing settings.

Limitations and More Work to Come
This study was done in controlled settings; more work needs to be done to make sure the results can be used in the real world. Adding AI to deal with constantly changing conditions could make the local reduction method work better in more complicated situations.